\documentclass[preprint,12pt]{elsarticle} 

\usepackage{graphicx}
\usepackage{amssymb}
\usepackage{lineno}

\usepackage{hyperref}[red]
\usepackage{longtable}
\usepackage{pdflscape}
\usepackage{multirow, booktabs}
\usepackage{multicol}
\usepackage{amsmath}
\usepackage{tikz} 
\usepackage{graphicx}
\usetikzlibrary{positioning}
\usepackage[utf8]{inputenc}
\usepackage{graphicx}
\usepackage{booktabs}
\usepackage{amsthm}
\usepackage{amsmath}
\usepackage{amssymb}
\usepackage{hyperref}[red]
\usepackage{amsfonts}
\usepackage{cuted,float}
\usepackage{makeidx}
\usepackage{amsthm,amscd,mathrsfs,latexsym,amsfonts,amssymb}

\usepackage[left=2cm,right=2cm,top=2cm,bottom=2.5cm]{geometry}
\usepackage{lipsum}
\usepackage{pagecolor}
\usepackage{afterpage}
\theoremstyle{plain}

\journal{Epidemics}

\begin{document}

\begin{frontmatter}


\title{Predicting malaria dynamics in Burundi using deep Learning Models}

\author[1]{Daxelle Sakubu\corref{mycorrespondingauthor}} \cortext[mycorrespondingauthor]{Corresponding author}
\ead{daxelle.sakubu@ub.edu.bi}
\author[1,2]{Kelly Joelle Gatore Sinigirira}
\author[1,2,3]{David Niyukuri}

\address[1]{Doctoral School, University of Burundi, Burundi}
\address[2]{Department of Mathematics, University of Burundi, Burundi}
\address[3]{The South African Department of Science and Technology--National Research Foundation (DST-NRF) Centre of Excellence in Epidemiological Modelling and Analysis (SACEMA), Stellenbosch University, Cape Town, South Africa}

\begin{abstract}
Malaria continues to be a major public health problem on the African continent, particularly in Sub-Saharan Africa. Nonetheless, efforts are ongoing, and significant progress has been made. In Burundi, malaria is among the main public health concerns. In the literature, there are limited prediction models for Burundi. We know that such tools are much needed for interventions design. In our study, we built machine-learning based models to estimates malaria cases in Burundi. The forecast of malaria cases was carried out at province level and national scale as well. Long short term memory (LSTM) model, a type of deep learning model has been used to achieve best results using climate-change related factors such as temperature, rainfal, and relative humidity, together with malaria historical data and  human population. With this model, the results showed that at country level different tuning of parameters can be used in order to determine the minimum and maximum expected malaria cases. The univariate version of that model (LSTM) which learns from previous dynamics of malaria cases give more precise estimates at province-level, but both models have same trends overall at provnce-level and country-level.

\end{abstract}


\end{frontmatter}


\section{Introduction}
 
Malaria is an infectious disease caused by the falciparum parasite, which is transmitted through the bite of a female mosquito named Anopheles. Malaria parasite are principally classified into four type: Plasmodium Falciparum, Plasmodium Vivax, Plasmodium Malaria, and Plasmodium Ovale. Plasmodium Falciparum is the most dangerous parasite since its incubation period is short 6-14 days\cite {sajana2017machine} and it's the principal cause of Malaria in Burundi nearly 90 \% \cite{mbb2007}. 

According to the World Health Organization (WHO) report \cite{ world2021world}, estimated malaria cases increased from 213 million to 228 million, and deaths increased from 534 000 to 602 000 in the WHO African Region between 2019 and 2020. This region accounted for approximately 95 \%  of all cases and 96 \% of all deaths globally, children under the age of five account for 80 \% of all deaths in this region.  

In Burundi, malaria infection is among main major public health concerns after  Diarrheal diseases, Neonatal disorders, and tuberculosis \cite{world2016world, world2017world, world2021world, ihme2020}. During the recent malaria outbreak in 2017 \cite{world2017weekly27, unicefBDI2017, lok2019malaria}, the World Vision International assumed that climate change, population density, changed agricultural methods, food hardship, and a lack of information and action to prevent malaria were the main driving factors that contribute to the disease's prevalence in Burundi \cite {wvi2017}.

On a global scale, coordinated initiatives to minimize the malaria epidemic are being planned as part of the millennium development goals. In Burundi, the package of interventions for malaria control during pregnancy comprises ITN promotion and use, IPTp with sulfadoxine-pyrimethamine under directly observed treatment, and quick and successful treatment. The NMCP has yet to follow WHO recommendations from 2016, which increased the recommended number of prenatal care visits from four to eight. With Malaria Operational Plan (MOP) funds from Fiscal Year (FY) 2021 \cite{mop2022}, the team proposes to trial an evidence-based group prenatal care strategy to enhance IPTp uptake (according to DHIS2, 54 \% of women received the recommended three doses at U.S. Government-supported health centers in the first quarter of FY 2021). Although massive effort have already been deployed, Malaria still prevail. Thus this project aim to understand the climatic contributions by using machine learning models to predict Malaria cases.

In recent years, there has been a great deal of interest in the development and application of machine learning (ML) in the field of infectious diseases \cite {wiemken2019machine}. Not only as a catalyst for academic studies, but also as a critical means of detecting pathogenic microorganisms, implementing public health surveillance, investigating host-pathogen interactions, discovering drug and vaccine candidates, and so on. According to one survey, ML is used in 77\% of the products we use today. Machine learning (ML) is a subfield of Artificial Intelligence (AI) that is an important tool in bioinformatics \cite {larranaga2006machine}. When confronted with a range of large and complex data sets that must be analyzed, ML may employ sophisticated algorithms and efficient models to extract meaningful information from vast amounts of complex data-sets \cite{chekol2018employing, masinde2020africa,nkiruka2021prediction} . 

Machine learning extracts useful information from enormous amounts of data by using algorithms to recognize patterns and learn in an iterative process. Instead of depending on any preconceived equation that may serve as a model, ML algorithms use computing methods to learn directly from data \cite {sajana2018ensemble}. The union of mathematics and computer science in ML has shown significant potential as a breakthrough in science and technology, and it has been applied to a wide range of scientific fields, including biology. In the case of Malaria prediction in Burundi some studies have already been done, and one has used machine learning such as artificial neural network for prediction \cite {mfisimana2022exploring}.The authors investigated malaria across different group of ages and the impact of meteorological factors on the high number of malaria cases during some seasons.

Different degrees of accuracy were reported from previous investigations. The overarching goal of this research is to investigate  malaria cases predictions  based on meteorological data. Deep learning model will be used to forecast malaria incidence in all five provinces. This study may help in the improvement of public health measures especially on district level. 

\section{Data Description} 

The study was carried out using monthly data, collected from different sources, namely: Geographical Institute of Burundi (\textit{Institut Géographique du Burundi, IGEBU)}, the Institute of Statistics and Economic Studies of Burundi \textit{Institut de Statistiques et d’Études Économiques du Burundi, ISTEEBU)} and Burundi National Malaria Control Program (NMCP). The data was collected for all the eighteen provinces of the country.

\subsection{Data extraction}

Data collected from IGEBU was on a monthly scale from 2010 to 2022 with parameters such as relative humidity, rainfall and temperature with their maximum and minimum values. The average was calculated and inserted in the data-set. Malaria historical data was obtained at NMCP Burundi on a monthly scale from January 2010 to December 2022 for all previously eighteen provinces. The human population feature was available online on ISTEEBU website and the human population was calculated annually. 

\subsection{Data processing}

The data collected was on different time scale monthly and annually, thus the human population was considered as constant during the whole year. The meteorological data contained some missing values that were filled using an algorithm called missForest \cite{stekhoven2015missforest}. This algorithm was judged to be suitable since it takes into account the possible relation between variables. Recently, the communes, zones and hills/neighbourhoods of the Republic of Burundi have been the subject of administrative redistricting-regrouping. Bujumbura (Western part), Buhumuza (Eastern part), Gitega (Central part), Burunga (southern part) and Butanyerera (Northern part) are the five provinces of the country in the new delimitation of provinces. 

In this reform, the provinces were regrouped as follow: Bujumbura included all the commune of Bujumbura Mairie,Bujumbura Rural, Bubanza and Cibitoke. Gitega assembled all the commune of Gitega, Mwaro, Karuzi and Muramvya. Buhumuza grouped together all the communes of Cankuzo, Muyinga and Ruyigi. Butanyerera aggregated all the communes of Kirundo, Ngozi and Kayanza. Finally, Burunga included all the communes of Bururi, Makamba, Rumonge and Rutana. Therefore ,the data-set of these new delimited provinces were the mean of the old provinces regrouped per month for the meteorological data and the sum for the human population and malaria cases data. 

\section{Methodology} 

After processing the data, we start building the neural network models.The prediction model which was build with four layers and several units. Nevertheless, the performance were not good, hence a different approach  was taken in order to find good results.

\subsection{Building ML models} 

 Since the data-set  were on a monthly scale, a different model that took into account the previous information seems to be appropriate for this study. The kind of algorithm that take into account previous information into a chronological manner are recurrent neural network. This kind of neural network in contrary to others have feedback connections. The network's connection weights and biases change once per training episode, similar to how physiological changes in synaptic strengths store long-term memories; the network's activation patterns change once per time-step, similar to how the brain's electrical firing patterns change moment to moment to store short-term memories. Recurrent neural networks are frequently used in multiple domain to predict future events based on prior experience. 
 
 For instance, LSTM can be used for tasks like connected, unsegmented handwriting recognition, video games,speech recognition,automated translation, healthcare, Speech activity detection and Robotics. Since LSTM has proven to have a good performance to aid with rational decision-making, it was utilized for prediction of malaria cases in this study. 
 
\begin{figure}[h!]
\begin{center} 
\includegraphics[scale=0.45]{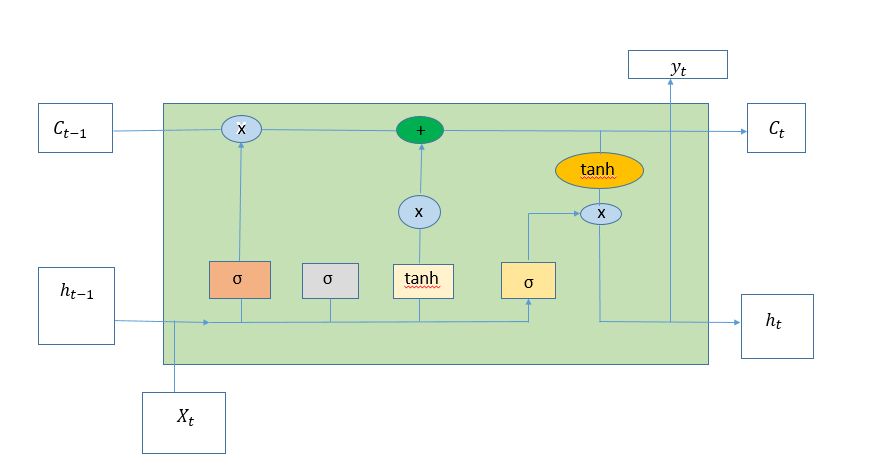}  
\caption{ LSTM cell}\label{fig1:LSTM cell}   
\end{center}
\end{figure}

In Figure \ref{fig1:LSTM cell}, a long short term memory cell is depicted where the square represent the layers , the ellipse are the component wise operation,Ci are hidden state vectors and Xi are input vector to the LSTM unit and hi are hidden state vector also known as output vector (yi) of the LSTM unit.

\begin{figure}[h!]
\begin{center} 
\includegraphics[scale=0.5]{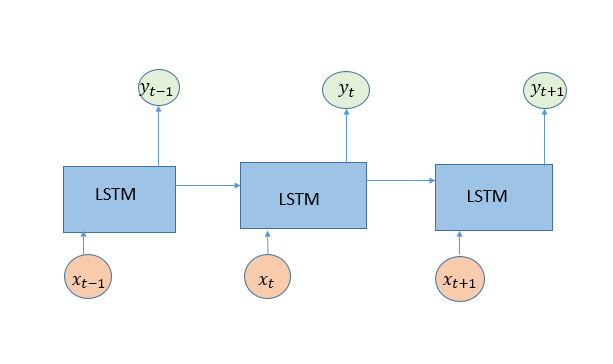}  
\caption{Univariate LSTM.}\label{fig2:Univariate LSTM}  
\end{center}
\end{figure}

In Figure \ref{fig2:Univariate LSTM}, a univariate LSTM model is shown in a sequential manner with previous states along side the input vectors $x_i$ and outputs $y_i$. The previous malaria cases are the input vector and the output vector are the actual malaria cases.

\begin{figure}[h!]
\begin{center} 
\includegraphics[scale=0.5]{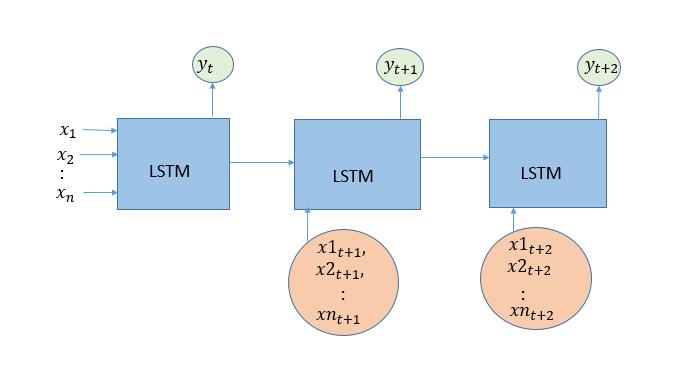}  
\caption{Multivariate LSTM.}\label{fig3:multivariate LSTM}
\end{center}
\end{figure}

The Multivariate LSTM model is represented in Figure \ref{fig3:multivariate LSTM} with  the multiple input vector $x_i$ up to $x_n$ and the target $y_i$. In the multivariate LSTM model the input vectors are the climate data human population and previous malaria cases, the output vector are the current malaria cases. 

After tuning the data-set in the ML models we obtained results which are presented in the following section.

\section{Results}   

After running and fitting the data to the models, the error between the actual and predicted cases were calculated using root mean square error(RMSE). From the obtained results are observed in Table \ref{tab:rmse}, overall,the univariate LSTM model got the smallest RMSE while the Multivariate got the biggest RMSE. In Bujumbura province the multivariate error is near four times as big as the univariate error. In the case of Gitega province the error gap is three times big between multivariate and univariate and this is the same for Butanyerera province as well. In Burunga province the multivariate error is twice bigger the univariate error. For Buhumuza province the error gap was huge, the multivariate error was about eight times bigger than the univariate one. On the country level prediction the multivariate error disparity was three times big the univariate one.

\begin{table}[h!]  
\centering
 \caption{RMSE of malaria cases between Univariate LSTM and Multivariate LSTM.}\label{tab:rmse}
 \begin{tabular}{l|l|l}
 \hline
 \textbf{Province} & \textbf{Univariate LSTM} & \textbf{Multivariate LSTM} \\ \hline  
 \textbf{Bujumbura} & $4868.69$ & $16777.17$ \\ 
    \textbf{Gitega} & $10943.18$ & $31012.00$ \\  
         \textbf {Burunga}& $6403.33$ & $12964.48$ \\  
          \textbf{Butanyerera} & $8288.07$ & $25187.09$ \\  
           \textbf{Buhumuza} & $5664.18$ & $44893.25$ \\  
              \hline 
            \textbf{Country level: Burundi} & $31635.95$ & $119724.68$ \\ 
               \hline
 \end{tabular} 
\end{table}

\subsection{Province-level predictions}

The experiment were done on eighty percent of the data-set while the testing was done on the rest of the data-set. Malaria cases prediction on province level are shown below in figure 5 from 2020 september to september 2022. In univariate LSTM prediction, the curve trend are followed in most cases with the observed cases being sligtly higher than the expected ones. In the multivariate LSTM prediction, the curve trend are not coordinated except for the country level prediction. For Gitega province the overall malaria cases during that period of time are 2395666, the univariate model predicted 2444657,664 while the multivariate predicted 2338582,423. In Burunga province the observed malaria cases were 1901832 while the univariate predicted were 1890808,86 and the multivariate 1974578,2. In Buhumuza province, the observed cases during that time was 2772583 whereas the Univariate model predicted 2808850,717 and the multivariate 3567723,29. For Butanyerera, the actual cases were 3568702 when the univariate predicted 3507692,91 and the multivariate 3808420,97. The Bujumbura province observed malaria cases were 2253588 during that amount of time when the expected was 2232670,579 for Univariate LSTM model and 2280516,361 for the multivariate LSTM model. 

\begin{figure}[h!]
\setkeys{Gin}{width=0.5\linewidth}
\includegraphics{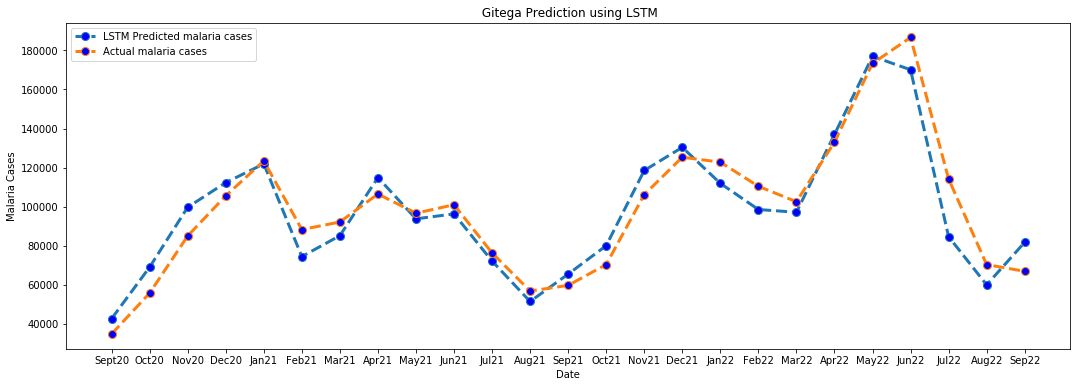}  
\hfill
\includegraphics{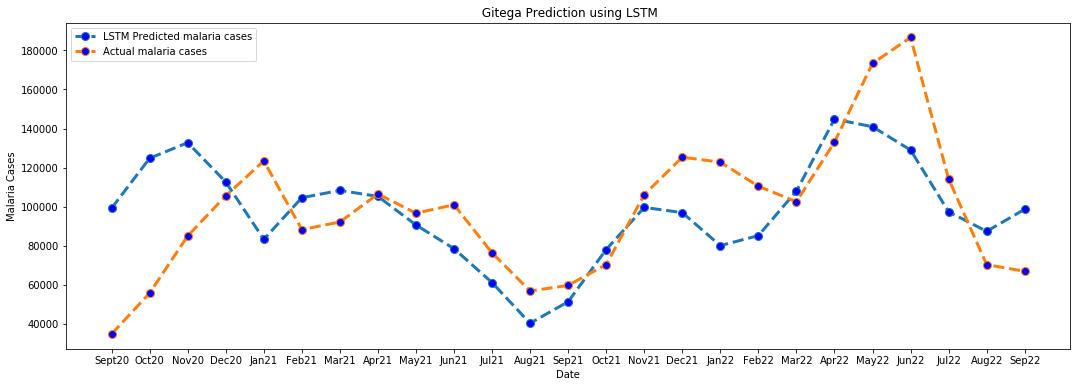}  
\smallskip
\includegraphics{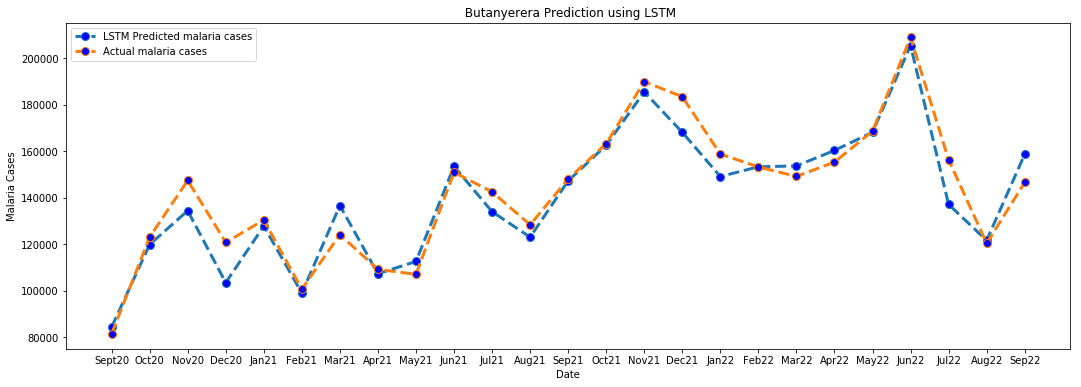}
\hfill
\includegraphics{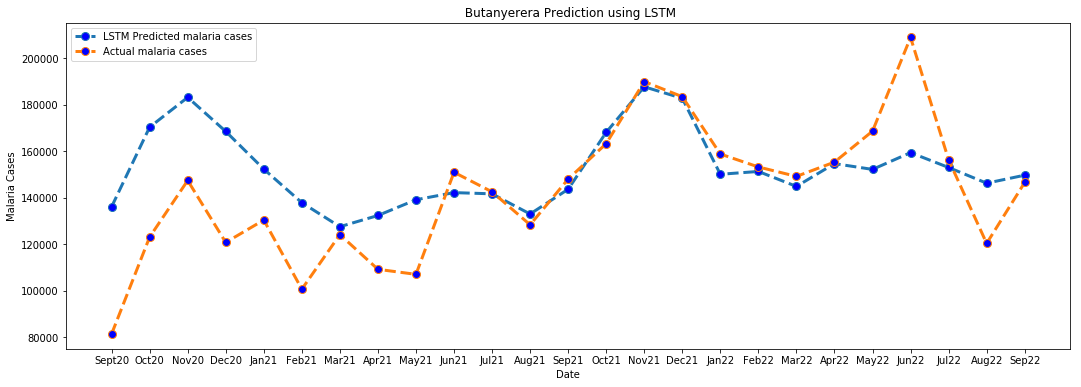} 
\smallskip
\includegraphics{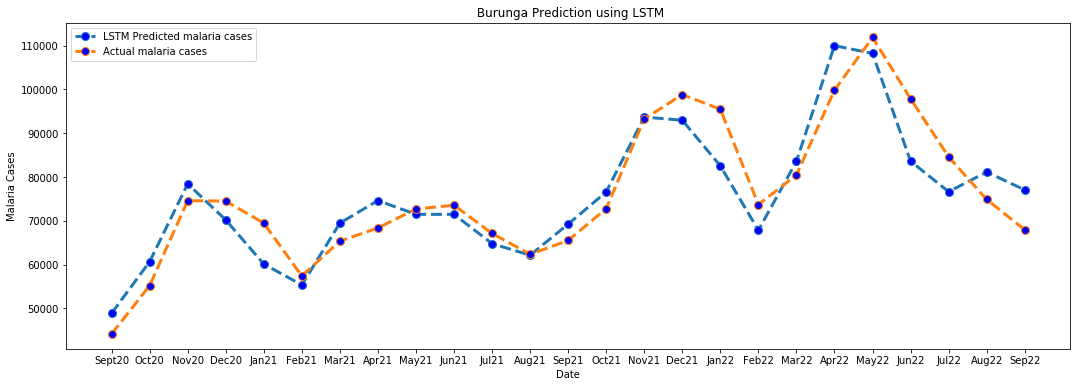}
\hfill
\includegraphics{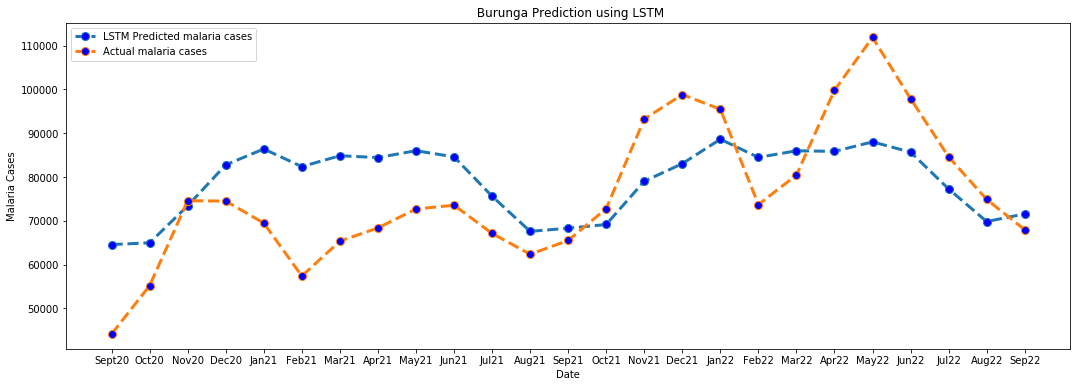} 
\smallskip
\includegraphics{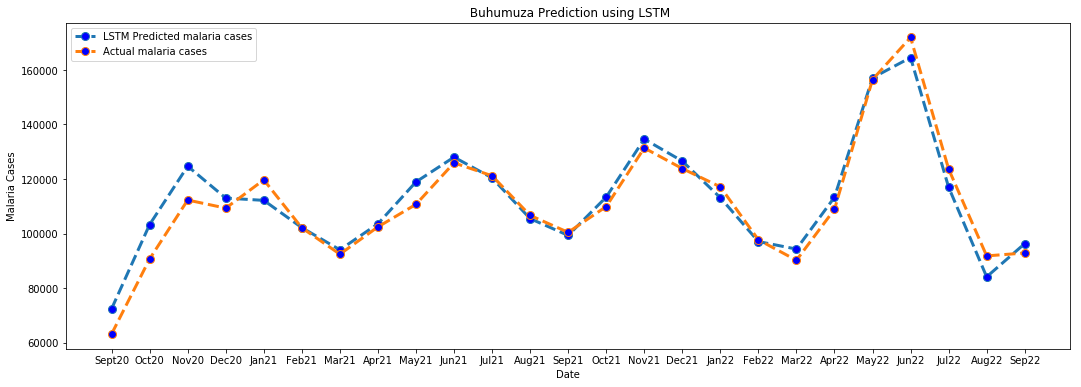} 
\hfill
\includegraphics{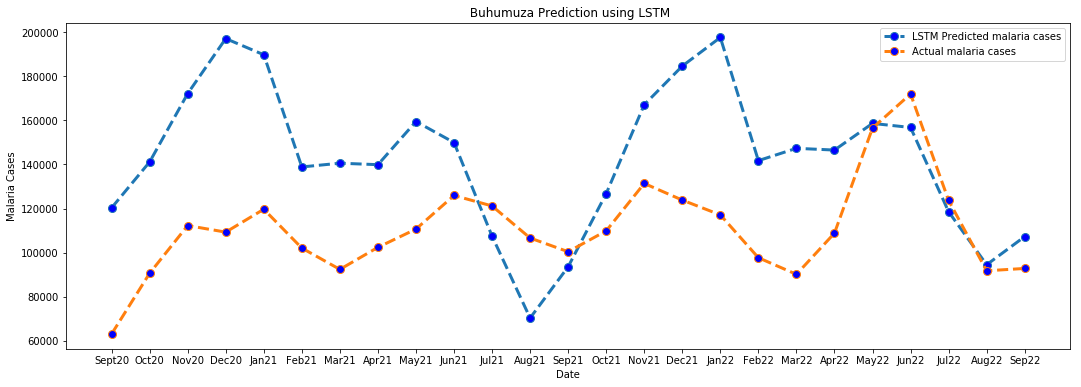}
\smallskip
\includegraphics{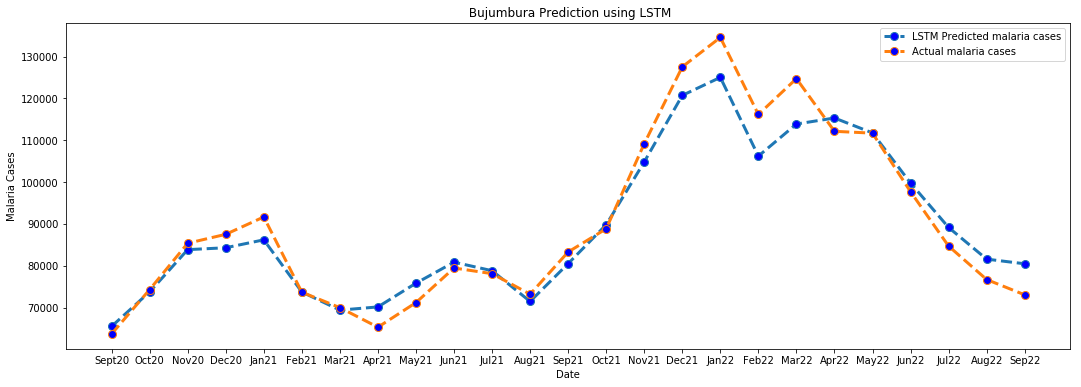} 
\hfill
\includegraphics{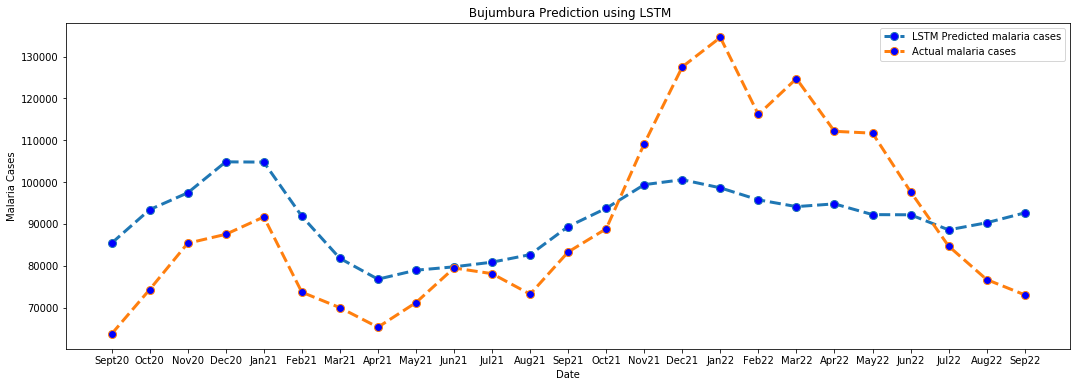}
\caption{Predictions of malaria cases in provinces using Univariate LSTM (left) and Multivariate LSTM (right)}
\label{tab:traintestsplit}
\end{figure}

\subsection{Country-level predictions}

After predicting malaria cases on province level, the country level prediction was done. The results are depicted in Figure \ref{tab:traintestsplit}.The data-set of the country level was the mean of all provinces for the climate data and the sum for the human population and malaria cases. The curve trends were coordinated to the actual trend in all models. Neverthless, the multivariate predicted more malaria cases than observed one while the univariate predicted less. The malaria cases that occurred in the country during the two years were 12959182,46, the univariate model predicted 12841653,9 which is around one million cases less than what was observed. The multivariate LSTM model predicted 15215766,15
that is around more than two millions and half than the actual cases.

\begin{figure}[h!]
\setkeys{Gin}{width=0.5\linewidth}
\includegraphics{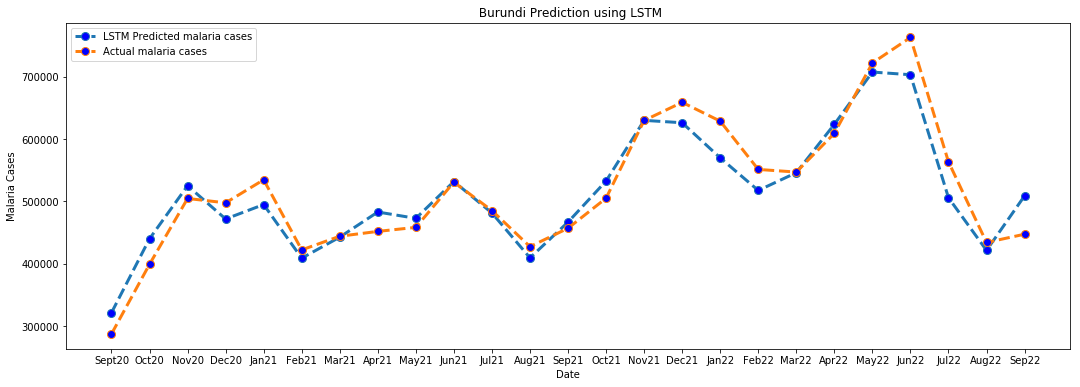}
\hfill
\includegraphics{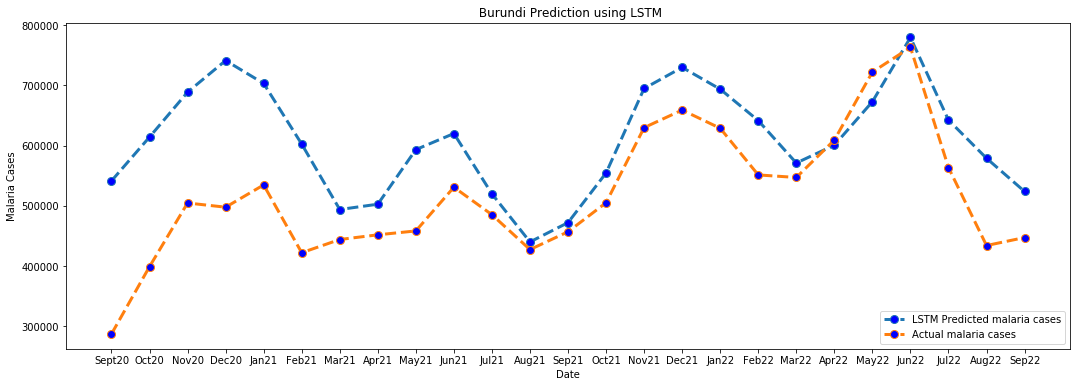}  
\caption{Predictions of malaria cases on country level using Univariate LSTM (left) and Multivariate LSTM (right)}
\label{tab:traintestsplit}
\end{figure}

\section{Discussion}

Based on the climate related variables, the LSTM model provided promising results on the prediction of malaria cases at country level. At province level, for the north-eastern part Butanyerera and Buhumuza, the multivariate LSTM tend to predict much more than observed. However, for Bujumbura and Burunga, the south-Eastern part , the trend were followed up until october 2021 and start to decrease in 2022. In Gitega province the central part, the expected cases from multivariate LSTM model were highly inferior to the observed cases. The univariate LSTM on the other hand gave the best accuracy with the curve trend being followed on country level as well as on the province level. 

This model proved to be useful in prediction of possible upcoming outbreaks. However, few points need to be underligned. Since the the models use previous information in order to predict the upcoming event, information gathering and sharing is critical, without the most update information the accuracy will not be the same and may decrease.

From Figure \ref{tab:traintestsplit}, we have seen that the Univariate LSTM results depict the overall trends of malaria cases, and it gives more precise estimates at province-level. With Univariate LSTM getting the best accuracy, multivariate LSTM may give the maximum cases that can be expected overall. The models got the same trends overall, especially on a country level prediction, this get to confirm how much the climate condition affect the malaria outbreak  in the country.


The effects of climate change on disease outbreaks go beyond malaria transmission dynamics. Investigating the meteorological factors in the case of malaria alongside historical data such as human population and total malaria cases per month was done in this work. With the results obtained, it's clear that despite artificial neural network acting like black box,there is a room for meteorological data being an important factor to be considered in the prediction of an outbreak especially on the country level where the maximum cases were predicted. The use of deep learning model such as recurrent neural networks has provided promising results in the prediction of malaria which still causes outbreak in sub-saharan Africa. More applications of artificial intelligence will help the collaborations between different intervention teams in the healthcare system, especially on how they should respond after predicting potential increase or decrease of malaria cases.   
\bibliographystyle{unsrt}
\bibliography{malar.bib}







\end{document}